\documentclass[conference]{IEEEtran}
\IEEEoverridecommandlockouts
\usepackage{amsmath}
\usepackage{amssymb}
\usepackage{balance}
\usepackage{booktabs}
\usepackage[noadjust]{cite}
\usepackage{graphicx}
\usepackage{siunitx}
\usepackage[absolute,showboxes]{textpos}
\usepackage{xcolor}
\usepackage[caption=false, font=footnotesize, subrefformat=parens, labelformat=parens]{subfig}
\graphicspath{{./img/}}
\TPMargin*{3pt}

\newcommand\copyrighttext{
	\footnotesize
	\noindent
	\textcopyright\,2023 IEEE.
	Personal use of this material is permitted.
	Permission from IEEE must be obtained for all other uses, in any current or future media, including reprinting/republishing this material for advertising or promotional purposes, creating new collective works, for resale or redistribution to servers or lists, or reuse of any copyrighted component of this work in other works.}%
\newcommand\copyrightnotice{%
	\begin{textblock*}{7in}(0.75in,0.25in)
		\copyrighttext
	\end{textblock*}
}

\begin{document}

\title{%
Automatic Intersection Management in\\Mixed Traffic Using Reinforcement Learning\\and Graph Neural Networks%
\thanks{Part of this work was financially supported by the Federal Ministry for Economic Affairs and Climate Action of Germany within the program "Highly and Fully Automated Driving in Demanding Driving Situations" (project LUKAS, grant numbers 19A20004A and 19A20004F).}%
}

\author{%
\IEEEauthorblockN{%
Marvin~Klimke\IEEEauthorrefmark{1}\IEEEauthorrefmark{2},
Benjamin~V\"olz\IEEEauthorrefmark{1}, and
Michael~Buchholz\IEEEauthorrefmark{2}
}\\
\IEEEauthorblockA{%
\IEEEauthorrefmark{1}Robert Bosch GmbH, Corporate Research, D-71272 Renningen, Germany.\\E-Mail: {\tt\small\{marvin.klimke, benjamin.voelz\}@de.bosch.com}\\
\IEEEauthorrefmark{2}Institute of Measurement, Control and Microtechnology, Ulm University, D-89081 Ulm, Germany.\\E-Mail: {\tt\small michael.buchholz@uni-ulm.de}%
}}

\maketitle
\copyrightnotice
\bstctlcite{ieeenodash}

\begin{abstract}
Connected automated driving has the potential to significantly improve urban traffic efficiency, e.g., by alleviating issues due to occlusion.
Cooperative behavior planning can be employed to jointly optimize the motion of multiple vehicles.
Most existing approaches to automatic intersection management, however, only consider fully automated traffic.
In practice, mixed traffic, i.e., the simultaneous road usage by automated and human-driven vehicles, will be prevalent.
The present work proposes to leverage reinforcement learning and a graph-based scene representation for cooperative multi-agent planning.
We build upon our previous works that showed the applicability of such machine learning methods to fully automated traffic.
The scene representation is extended for mixed traffic and considers uncertainty in the human drivers' intentions.
In the simulation-based evaluation, we model measurement uncertainties through noise processes that are tuned using real-world data.
The paper evaluates the proposed method against an enhanced first in - first out scheme, our baseline for mixed traffic management.
With increasing share of automated vehicles, the learned planner significantly increases the vehicle throughput and reduces the delay due to interaction.
Non-automated vehicles benefit virtually alike.
\end{abstract}

\section{Introduction}
\label{sec:intro}

Traffic congestion is a major cause of time loss and energy inefficiency in urban mobility.
Ever increasing traffic demands challenge the currently prevailing approaches for managing intersections, like static priority rules or simple traffic light schemes.
The issue is aggravated by occlusion, e.g., through buildings and other road users, which limits the view onto cross traffic for human drivers as well as on-board perception systems.

The deployment of connected automated driving has the potential to alleviate many of these issues.
Thereby, human driven connected vehicles (CVs) and connected automated vehicles (CAVs) share a communication link with each other and possibly infrastructure systems.
By providing edge computing resources in urban areas, an environment model of the local traffic scene, for instance at an intersection, can be maintained and provided to connected road users.
This information could be used, e.g., by CAVs to automatically merge into a gap in prioritized traffic, as shown in \cite{buchholz2021handling}.

\begin{figure}
	\centering
	\includegraphics[width=\linewidth]{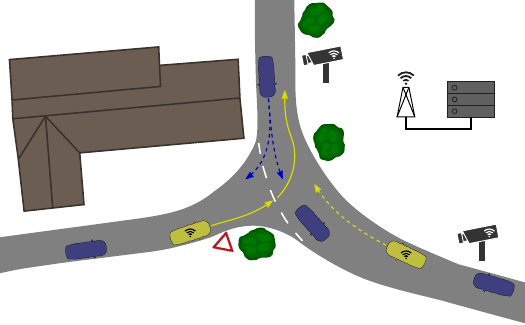}
	\caption{Cooperative planning in mixed traffic. The two automated vehicles (depicted in yellow) may perform a cooperative maneuver by deviating from the precedence rules, if it does not interfere with regular vehicles. Here, the maneuver of the regular vehicle (in blue) in the top has to be considered.}
	\label{fig:intro}
\end{figure}

Further increase in traffic efficiency can be obtained by leveraging automatic intersection management (AIM).
For example, an automated vehicle on the main road may be requested to slow down and let another vehicle coming from a side road merge, as illustrated in Fig.~\ref{fig:intro}.
On public roads, CAV penetration will not be even close to \SI{100}{\percent} soon, but mixed traffic will be prevalent.
Thus, cooperative planning approaches have to take all vehicles into account, including those that cannot be influenced.

In the current work, we propose a mixed-traffic-capable cooperative behavior planning scheme using reinforcement learning (RL) and graph neural networks (GNNs).
We build our approach on our previous work proposing a cooperative planning model for fully automated traffic \cite{klimke2022cooperative,klimke2022enhanced}.
The contribution of the current work is fourfold:
\begin{itemize}
\item To the best of our knowledge, we propose the first learning-based AIM scheme for mixed traffic,
\item Introduction of a novel GNN architecture leveraging an attention mechanism and relation-dependent edges,
\item Presenting an RL training procedure taking into account uncertainty about non-connected vehicles,
\item Demonstration of the model's ability to increase traffic efficiency over varying automation rate.
\end{itemize}

The remainder of the paper is structured as follows:
Section~\ref{sec:sota} gives an overview on the state of the art in AIM and machine learning based planning for automated driving.
We present our learning model as well as details on the graph structure in Section~\ref{sec:approach}.
The experimental simulation setup and evaluation results are discussed in Section~\ref{sec:eval}.
The paper is concluded in Section~\ref{sec:conclusion}.

\section{Related Work}
\label{sec:sota}

The state of the art in AIM consists of a broad variety of approaches, many of which can be allocated to reservation-based systems \cite{dresner2004multiagent,dresner2008multiagent,nichting2019explicit,nichting2020space}, optimization algorithms \cite{kamal2015vehicle-intersection,malikopoulos2018decentralized,hult2018miqp-based,li2018near-optimal}, or tree-based methods \cite{kurzer2018decentralized,mertens2022cooperative}.
In the following, we discuss a selection of works in commonly used paradigms and refer the reader to surveys like \cite{zhong2020autonomous} for a more extensive overview.

The tile-based reservation scheme proposed in \cite{dresner2004multiagent} employs a central coordination unit that assigns clearance to the vehicles in order of their requests.
This first in - first out (FIFO) policy is benchmarked on simulated multi-lane intersections against traffic lights as a baseline and an overpass as the ideal solution while disallowing turning maneuvers.
In \cite{dresner2008multiagent}, the approach is extended to support turns at the intersection and traditional yielding as an additional baseline.
This work also investigates intersection management in mixed traffic, however, requires strict spatial separation between human-driven vehicles and CAVs.

Connected automated driving also supports ad-hoc negotiation of cooperative maneuvers \cite{nichting2019explicit,nichting2020space}.
In \cite{nichting2019explicit}, a cooperative lane change procedure is presented.
If a CAV deems a lane change maneuver beneficial, it may request surrounding vehicles to keep a designated area free to safely perform the lane change.
The authors extend their approach for usage at intersections by introducing a cooperation request when planned paths conflict on the intersection area \cite{nichting2020space}.
The feasibility of this procedure is demonstrated in low-density real traffic using two testing vehicles.

In fully automated traffic, intersection efficiency can be further improved by optimizing the individual vehicles' trajectories freely on the 2D ground plane.
To conquer the sharply rising complexity with increasing number of agents, standard cases can be defined and solved offline, as proposed by \cite{li2018near-optimal}.
Prerequisite for applying the precomputed solutions to a given traffic scene online, is to have the approaching vehicles form a predefined formation before entering the intersection area.
The authors acknowledge that high computing power is needed to solve all standard cases for a multi-lane four-way intersection, even when done offline.
Other approaches assume the vehicles to follow predefined lanes and solely optimize the longitudinal motion \cite{kamal2015vehicle-intersection,malikopoulos2018decentralized,hult2018miqp-based}.
Centralized model predictive control of the longitudinal vehicle motion is employed by \cite{kamal2015vehicle-intersection}.
In \cite{malikopoulos2018decentralized}, a distributed energy-optimizing approach is presented that disallows turning maneuvers.
Both works demonstrate a reduction in delay and fuel consumption compared to traditional signalized intersections.
Coordination by optimizing the intersection crossing order using mixed-integer quadratic programming can yield a benefit over FIFO ordering when considering the increased inertia of heavy vehicles \cite{hult2018miqp-based}.
These approaches are not suited for mixed traffic and share the unfavorable scaling of computational demand with increasing number of road users.

Decision trees are well-suited to represent possible variants of the scenario evolution depending on agent actions or interactions.
In \cite{kurzer2018decentralized}, Monte Carlo tree search is employed for decentralized cooperative trajectory planning in highway scenarios.
The authors demonstrate their model's ability to generate viable trajectories in an overtaking maneuver for varying levels of cooperation.
In the context of AIM, decision trees enhanced by probabilistic predictions of scene evolvement can be used to find the crossing order, which yields the highest expected efficiency \cite{mertens2022cooperative} in mixed traffic scenarios.

In recent works, machine learning demonstrated its potential for automated driving, yielding remarkable results in road user motion prediction \cite{lefevre2014survey} and planning for a single ego vehicle \cite{zhu2021survey}.
Many planning models \cite{chen2019deep,bansal2019chauffeurnet} rely on the imitation of expert demonstrations taken, e.g., from driving datasets using supervised learning.
In \cite{cui2022coopernaut}, it is proposed to aid a machine learning model for single ego planning using inter-vehicle communication providing LiDAR measurements from surrounding vehicles to enhance the sensor coverage.
For cooperative behavior planning, ground truth data is virtually unavailable, because such maneuvers are seldom, if at all, performed by human drivers.
Therefore, supervised learning approaches are rather limited for the application in AIM.
Reinforcement learning (RL) evades the need for large amounts of training data, by instead exploring possible behaviors in simulation, guided by a reward signal.
The application of RL to single ego planning was demonstrated for handling urban intersections \cite{capasso2021end--end} and highway lane change maneuvers \cite{hart2020graph}.
The authors of the latter work propose a graph-based scene representation for encoding the ego vehicle's semantic surroundings and a GNN-based policy.

Learning-based algorithms have not been as extensively used for cooperative multi-agent behavior planning in automated driving.
Multi-agent reinforcement learning approaches have been proposed for a large variety of problem settings, e.g., in robotics.
Due to the specific nature of server-side cooperative planning and presence of human-driven vehicles, these works are not transferable to our task.
In \cite{wu2019dcl-aim}, RL is employed to train a policy for choosing the most suited action from a restricted discrete action space, which ensures collision-free maneuvers.
Because the policy execution is performed decentralized for each vehicle individually, the approach does not leverage explicit communication and dedicated cooperation between agents.
Our previous work \cite{klimke2022cooperative} was the first to propose a GNN trained in RL for centralized cooperative behavior planning in fully automated traffic.
The representation and learning model have been improved and was shown to generalize to intersection layouts not encountered during training \cite{klimke2022enhanced}.

\section{Proposed Approach}
\label{sec:approach}

This section introduces the proposed learning paradigm (Sec.~\ref{ssec:learningparadigm}), the graph-based input representation (Sec.~\ref{ssec:inputrepresentation}), the network architecture (Sec.~\ref{ssec:network}), and the reward function (Sec.~\ref{ssec:reward}).

\subsection{Learning Paradigm}
\label{ssec:learningparadigm}

The task of cooperative planning across multiple vehicles is considered as a multi-agent RL problem.
Depending on the degree of centralization, different learning paradigms can be employed \cite{gronauer2021multi-agent}.
Instead of deploying multiple agents independently within the simulated environment, joint cooperative behavior planning is best modeled by the centralized training centralized execution paradigm.
All decentralized approaches require the agents to develop an implicit communication scheme implemented through their behavior.
In connected automated driving, however, a communication link is at the disposal of the involved agents and makes centralized execution feasible.
Therefore, the multi-agent planning task is modeled as a single partially observable Markov decision process (POMDP), denoted as
\begin{equation}
(S,\,A,\,T,\,R,\,\Omega,\,O).
\label{eq:pomdp}
\end{equation}
The set of states $S$ contains all reachable states of the simulator, including full state information on all automated vehicles (AVs) and manually driven vehicles (MVs).
While the set of available actions is denoted by~$A$, the conditional transition probability of changing from $s \in S$ to $s' \in S$ when applying action $a \in A$ is given by $T(s'|s,\,a)$.
The state transition is not deterministic because human reaction to a given traffic scene is neither, which has to be modeled in simulation.
$R: S \times A \rightarrow \mathbb{R}$ denotes the reward function that rates the chosen action in a given state in terms of a scalar value.
Because most of the abstract traffic state $S$ is not observable for the multi-agent planner, a reduced set of observations is defined as $\Omega$.
The probability of a state $s \in S$ being mapped to observation $\omega \in \Omega$ is given by $O(\omega|s)$.
This mapping is not injective, which manifests, for instance, when the driver of a non-connected vehicle changes their turning intention without any externally noticeable change.
Moreover, measurement uncertainties are modeled by $O$.
The dimensionalities of the state space and the action space depend on the number of (controllable) vehicles currently in the scene and may vary over time.

\subsection{Input Representation}
\label{ssec:inputrepresentation}

\begin{figure}
	\centering
	\includegraphics[width=\linewidth]{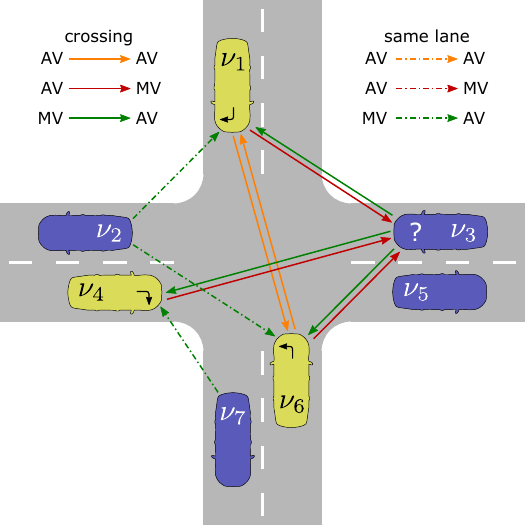}
	\caption{The graph-based input representation for mixed traffic at a four-way intersection. An AV's (yellow) turning intention is denoted by an arrow on its hood. Due to the unknown turning intention of the MV $\nu_3$ (blue, denoted by '?'), it shares edges with all three AVs, although conflicts with $\nu_1$ and $\nu_4$ are mutually exclusive.}
	\label{fig:scene_graph_mixed}
\end{figure}

We retain the core idea of encoding the traffic scene at an urban intersection as a directed graph with vertex features and edge features from~\cite{klimke2022enhanced}.
Graph-based input representations proved to be well-suited for behavior planning in automated driving, where a varying number of dynamically interacting entities must be encoded efficiently~\cite{huegle2020dynamic}.
The set of observations can thus be denoted as $\Omega = (V,\,E,\,U)$, where each vehicle is mapped to a vertex $\nu \in V$, $E$ describes the set of edges, and $U$ a set of edge types.
The available edge types result from the Cartesian product between the relation-dependent dimension and the automation-dependent dimension: $U = U_\text{rel} \times U_\text{aut}$.
While $U_\mathrm{rel} = \{\text{same~lane},\,\text{crossing}\}$ is retained from \cite{klimke2022cooperative}, $U_\mathrm{aut} = \{\text{AV/AV},\,\text{AV/MV},\,\text{MV/AV}\}$ denotes whether the edge encodes the interaction between two AVs, one AV and one MV or vice versa.
The reason for this design follows from the fact that the coordination of two AVs is fundamentally different than an interaction of an AV with an MV.
Note that there are no edges between two MVs, because pure MV interactions are not of concern for the cooperative planner.
In case of, e.g., two MVs leading an AV, both MVs are connected via edges to the AV, which enables the network to infer relevant AV interactions.
Formally, an edge is defined as
\begin{equation}
(\nu_i,\,\nu_j,\,g_{ij},\,r) \in E,
\label{eq:edge}
\end{equation}
where the source and destination vertices are named $\nu_i$ and $\nu_j$, respectively.
The edge feature $g_{ij}$ will be described below and $r \in U$ specifies the edge type.
Two graph vertices are being connected by an edge, if the corresponding vehicles share a conflict point on the intersection area (crossing) or are driving on the same path (same lane), as illustrated in Fig.~\ref{fig:scene_graph_mixed}.
Note that the different edge types are mutually exclusive.

A key challenge of planning in mixed traffic is the ambiguity on the maneuver intention of non-connected vehicles, i.e., they could go straight, turn left or turn right (e.g. $\nu_3$ in Fig.~\ref{fig:scene_graph_mixed}).
In the graph-based scene representation, this issue is addressed by including all potential conflict points by means of additional edges.
Thereby, the planner becomes more cautious, because each conflict point may result in a collision if not coordinated properly.
As soon as the future motion of an MV can be predicted with reasonable accuracy, the edges for all other options can be removed.
In the present work, we assume an MV's intention to be predictable when it passed \SI{25}{\percent} of the intersection area.
The cooperative planning performance might be increased further by employing a prediction algorithm like~\cite{strohbeck2020multiple}.
Such an extension is, however, out-of-scope for this work.

The sets of input features for vertices and edges are adapted to accommodate the advanced requirements for planning in mixed traffic.
The vertex input features are denoted as one four-element vector per vehicle
\begin{equation}
\boldsymbol{h}^{(0)} = [s,\,v,\,\tilde{a},\,c]^T,
\label{eq:vertexfeature}
\end{equation}
where the upper index $(0)$ denotes the input layer of the GNN.
The first three elements describe the longitudinal position along its lane, the scalar velocity, and measured acceleration, respectively.
$c$ is a binary indicator on whether the corresponding vehicle is controllable, i.e., is an AV.
The edge input feature with distance measure~$d_{ij}$ and heading-relative bearing~$\chi_{ij}$ is extended by the new feature~$pr_{ij}$:
\begin{equation}
\boldsymbol{g}_{ij}^{(0)} = [1/d_{ij},\,\chi_{ij},\,pr_{ij}]^T.
\label{eq:edgefeature}
\end{equation}
This new feature encodes the priority relation between the two vehicles and is defined as
\begin{equation}
pr_{ij} = \max\left( \min(pr_i - pr_j,\,1),\,-1 \right),
\label{eq:priority}
\end{equation}
where $pr_i$ is the priority of vehicle~$i$ depending on its originating road and turning intention, given as an integer value.
In case of an MV with uncertain maneuver intention, its true priority is replaced by the assumed priority, given the worst-case maneuver with respect to the interacting AV.

\subsection{Network Architecture}
\label{ssec:network}

In this work, we employ the TD3~\cite{fujimoto2018addressing} RL algorithm, which belongs to the family of actor-critic methods for actions in continuous space.
As the actor and critic network architecture deviate only slightly on the output side, only the actor network is presented in detail for conciseness.
We propose a GNN architecture that is composed of relational graph convolutional network (RGCN) layers~\cite{gangemi2018modeling} and graph attention (GAT) layers~\cite{velickovic2018graph}.
To process edge features in the RGCN layers, we use the extended update rule for message passing that was introduced in~\cite{klimke2022enhanced}.
The different edge types correspond to independently learnable weight matrices.
Thus, the network can map the fundamental difference in interaction, as described in Sec.~\ref{ssec:inputrepresentation}, to sensible actions.

\begin{figure}
	\centering
	\includegraphics[width=\linewidth]{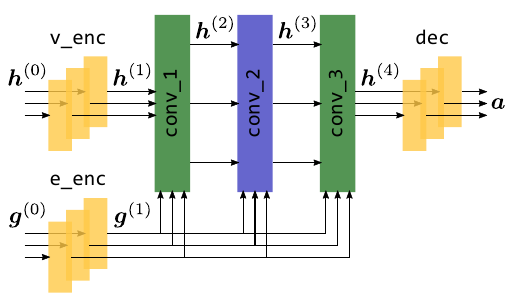}
	\caption{The GNN architecture of the actor network. Vertex input features $\boldsymbol{h}^{(0)}$ and edge input features $\boldsymbol{g}^{(0)}$ are mapped to one joint action $\boldsymbol{a}$. The edge feature enhanced RGCN layers are depicted in green, the GAT layer in blue, and fully connected layers in yellow.}
	\label{fig:actor}
\end{figure}

The overall network architecture is depicted in Fig.~\ref{fig:actor}.
Both the vertex input features and edge input features are first mapped into a high-dimensional space using the encoders \texttt{v\_enc} and \texttt{e\_enc}, respectively.
Afterwards, the vertex features in latent space are passed through three GNN layers for message passing, each of which gets the encoded edge features as an additional input.
All GNN layers fulfill the mapping $\mathtt{conv}: V^n \times E^m \rightarrow V^n$, where $n$ denotes the number of vehicles (nodes) and $m$ the number of conflict relations (edges).
The edge features are not updated throughout the forward pass, which is fine for our application, as there is nothing to be inferred on the edges.
Using alternating layer types proved to deliver better results than a pure RGCN or GAT network in our experiments.
Thus, we leverage the GAT layer's attention mechanism and explicitly include edge features and edge types in the modified RGCN layers.
The original GAT layer disregards edge type information and only considers the edge features for computing the attention weights but not the node update.
Recently, a graph attention mechanism for relational data was proposed~\cite{busbridge2019relational}, which did not perform well in our case, though.

The latent vertex features $\boldsymbol{h}^{(4)}$ are then passed to an action decoder, composed of fully connected layers, that infers a joint action for all vehicles.
Each intermediate layer is followed by a rectified linear unit (ReLU) as the activation function and all fully connected layers share their weights across vertices or edges.
The critic network aggregates the latent vertex features $\boldsymbol{h}^{(4)}$ to a single feature vector, which is then decoded into one q-value estimate.
The graph-based scene representation and the GNN are implemented using the PyTorch Geometric API~\cite{pyg}.

\subsection{Reward Function Engineering}
\label{ssec:reward}

The reward function for driving the RL training is composed of a weighted sum of reward components
\begin{equation}
R = \sum\limits_{k \in \mathcal{R}} w_k R_k \, ,
\end{equation}
where the weights are denoted as~$w_k$ and the set of reward components as~$\mathcal{R}$.
We retain the weights and reward components for velocity, action, idling, proximity, and collisions from~\cite{klimke2022cooperative}, with certain adaptions for mixed traffic described in the following.
To encourage cooperative maneuvers, the velocity reward not only considers the speed of AVs that are directly controlled by the RL~planner, but also MVs in equal weighting.
Therefore, a cooperative maneuver becomes even more attractive, if further MVs subsequently benefit from it.
The action penalty $R_\mathrm{action} = -||\boldsymbol{a}||_1$, on the other hand, only considers AVs to encourage smooth driving commands.

When applied in mixed traffic, this reward function exhibits an unintentional local optimum, resulting in AVs being stopped far away from the intersection entry and MVs subsequently passing the intersection without cross traffic.
This behavior is not desirable, as it effectively leads to a strong disturbance on the prioritized road whenever an MV appears on the minor road.
We propose to conquer this issue through an additional reluctance reward component
\begin{equation}
R_\mathrm{reluctance} = -\max\limits_{i \in \text{AVs}} \, \mathbb{I}(\nu_i \text{ is leader}) \; \mathbb{I}(v_i < v_\text{stop}) \; \delta_i,
\end{equation}
where $\mathbb{I}(\cdot)$ denotes the indicator function.
This penalty is only nonzero, if there is no leading vehicle and the velocity falls below a stopping threshold of $v_\text{stop} = \SI{1}{\meter\per\second}$.
$\delta_i$ denotes the distance of $\nu_i$ to the stop point in front of the intersection.
Throughout this study, the reluctance reward weight was set to $w_\mathrm{reluctance}=0.01$, based on empirical observations.

\section{Experiments}
\label{sec:eval}

In this section, the modeling of measurement uncertainties (Sec.~\ref{ssec:uncertainties}) and the simulation setup (Sec.~\ref{ssec:sim}) is introduced, before presenting evaluation results in Sec.~\ref{ssec:results}.

\subsection{Modeling of Measurement Uncertainties}
\label{ssec:uncertainties}

In addition to the epistemic uncertainty in the maneuver intention of non-connected vehicles, there is also aleatoric uncertainty due to measurement noise.
Although current testing vehicles might be equipped with a highly accurate GNSS-aided inertial navigation system (GNSS-INS), perfect localization cannot be assumed for the deployed system.
This issue becomes even more pressing with regular vehicles being present that must be localized via perception.
Therefore, we model measurement uncertainties in simulation by means of additive noise processes, whose parameters were estimated using real-world driving logs.
In this work, we use recordings of test drives of a connected testing vehicle at the pilot site in Ulm-Lehr, Germany~\cite{buchholz2021handling}, but the proposed method is applicable for larger amounts of data to obtain a more precise estimate.
During the test drive, the GNSS-INS track of the testing vehicle was recorded.
Additionally, the environment model state, which results from fusion of infrastructure sensor perception, was saved.

\begin{figure}
	\centering
	\includegraphics[width=\linewidth]{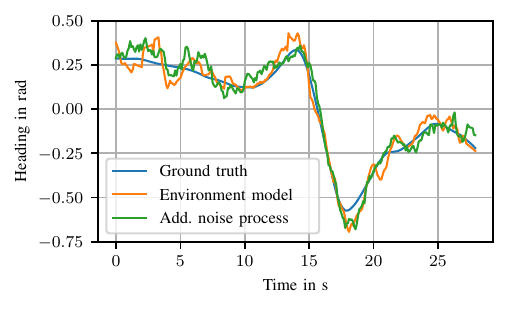}
	\caption{Exemplary result of superposing the measurement noise process to the vehicle heading during one traversal of the testing site. The qualitative nature of noise in the environment model's track is matched well by the modeled noise process.}
	\label{fig:noise_psi}
\end{figure}

We consider the GNSS-INS track as the ground truth to the environment model track and strive to parameterize four independent noise processes for the set of measured quantities $\lambda\in\{x,\,y,\,v,\,\psi\}$, being position in a local east-north-up frame, velocity, and heading.
The model deviation in each measurand is given as $e_\lambda = |\lambda - \tilde{\lambda}|$, where $\tilde{\lambda}$ denotes the environment model state that has been aligned temporally to the ground truth samples $\lambda$.
The sample frequency is~\SI{10}{\hertz}.
Temporal dependencies within the noise shall be modeled by a first-order autoregressive process (AR(1)), defined as
\begin{subequations}
\begin{equation}
\hat{e}_{\lambda,\,k} = \phi_\lambda\hat{e}_{\lambda,\,k-1} + \varepsilon_{\lambda,\,k},
\label{eq:noiseprocess}
\end{equation}
\begin{equation}
\varepsilon_{\lambda,\,k} \sim \mathrm{N}(0,\,\sigma_\lambda^2),\,\mathrm{i.i.d.},
\label{eq:addnoise}
\end{equation}
\end{subequations}
where $\phi_\lambda$ and $\sigma_\lambda^2$ are the process parameters for measurand~$\lambda$.
The subscript~$\lambda$ is dropped in the following for brevity.
We employ ordinary least squares estimation \cite{fox2016applied} to estimate the process parameter and its variance:
\begin{equation}
\hat{\phi} = \frac{\sum_{k=1}^{K} e_{k-1}e_k}{\sum_{k=1}^{K} x_{k-1}^2},
\label{eq:phi}
\end{equation}
\begin{equation}
\hat{\sigma}^2 = \frac{1}{K} \sum\limits_{k=1}^{K} \left(e_t - \hat{\phi}e_{k-1}\right)^2.
\end{equation}
Here, the number of samples in the recorded error track is denoted $K$.
The retrieved parameter set for our dataset is given in Table~\ref{tab:noiseparameters}.
Figure~\ref{fig:noise_psi} exemplarily shows the additive noise process for the vehicle heading compared to actual recordings of the environment model.
Note that, because of the stochastic nature of measurement uncertainties, it was not to be expected to observe an absolute match.
Instead, each run will yield different trajectories.

\begin{table}
	\caption{Noise Process Parameters}
	\label{tab:noiseparameters}
	\centering
	\begin{tabular}{lcccc}
		\toprule
		Parameter        & $x$  & $y$  & $v$  & $\psi$ \\ \toprule
		$\hat{\phi}$     & 0.968 & 0.968 & 0.957 & 0.936 \\ \midrule
		$\hat{\sigma}^2$ & \SI{0.014}{\meter\squared} & \SI{0.014}{\meter\squared} & \SI[per-mode=reciprocal]{0.045}{\meter\squared\per\second\squared} & \SI{0.0005}{\radian\squared} \\ \bottomrule
	\end{tabular}
\end{table}

\subsection{Training and Evaluation Environment}
\label{ssec:sim}

\begin{figure*}
	\centering
	\subfloat[eFIFO]{\includegraphics[width=0.5\textwidth]{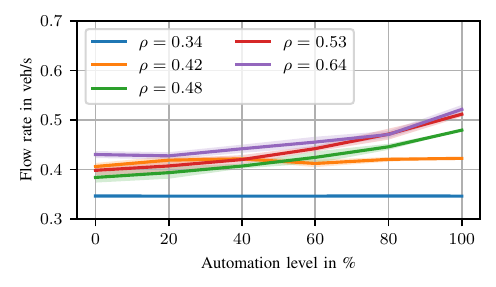}\label{sfig:flow_fifo}}%
	\subfloat[RL planner ($\times$ marks the results of the legacy model from~\cite{klimke2022enhanced})]{\includegraphics[width=0.5\textwidth]{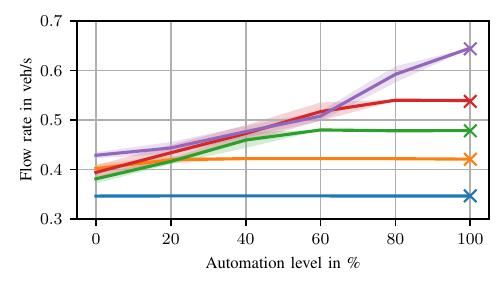}\label{sfig:flow_rl}}%
	\caption{Attained flow rate over varying automation level for the eFIFO baseline and the RL~planner. Each colored line represents one scenario definition with increasing traffic demand $\rho$ in veh/s. The shaded area indicates the standard deviation of the metric samples.}
	\label{fig:flowrate}
\end{figure*}

Training and evaluation of the proposed cooperative planning scheme was conducted in the open-source simulator Highway-env~\cite{highway-env} that was extended for usage in centralized multi-agent planning.
A kinematic bicycle model~\cite{kong2015kinematic} is applied for simulating plausible vehicle trajectories according to the action output of the RL algorithm.
The MVs' behavior models in Highway-env are based on car following models, like the intelligent driver model (IDM)~\cite{treiber2000congested} and have been tweaked for improved yielding behavior at intersections~\cite{klimke2022cooperative}.
For evaluation, the extended intelligent driver model (EIDM)~\cite{salles2020extending} is employed, which resembles human driving behavior more closely and features non-deterministic outputs that manifest in the transition $T$.
Unless noted otherwise, measurement noise according to the noise processes introduced in Sec.~\ref{ssec:uncertainties} is added in the observation $O$ during evaluation.

Training of the planner network begins in fully automated traffic for the first third, before the share of MVs is gradually increased during the second third and remains at \SI{50}{\percent} for the last third.
Beginning with \SI{100}{\percent} AVs allows the RL algorithm to learn basic behavior, like collision avoidance, before tuning the policy for the more challenging case of mixed traffic.
Instead starting the training with samples of solely MVs does not provide a benefit, because it lacks collision samples, thus preventing the RL algorithm from learning how to effectively avoid collisions.
During training, the idealized IDM car following model is employed, which yields superior results to using the EIDM.
Notably, this still holds for evaluation with the more diverse EIDM, which shows that the model does not overfit to the modeled human driving behavior.

In this work, we use an enhanced first in - first out (eFIFO) scheme as a baseline to the proposed RL~planner.
A more extensive set of baselines was employed in~\cite{klimke2022enhanced} for fully automated traffic, while we restrict the current analyses to the best performing baseline for conciseness.
The algorithmic idea of the eFIFO is being extended to handle mixed traffic.
Therefore, AVs are prioritized according to their current distance to the intersection.
Clearance to cross the intersection is assigned respecting the precedence relations to the MVs as boundary conditions, while conflict-free paths may be driven on simultaneously.

\subsection{Simulation Results}
\label{ssec:results}

We evaluate the planning approaches in five scenarios of varying traffic density at a four-way intersection and under different automation levels, i.e., proportion of AVs in traffic.
The simulated intersection connects a major road with a minor road that carries comparatively less traffic.
Each configuration was run ten times, because the non-determinism in the setup may cause volatility in closed-loop metrics.
In intersection management, the throughput of vehicles is of particular interest and is captured by the \emph{flow rate} metric in Fig.~\ref{fig:flowrate}.

Both planning approaches achieve a benefit for high automation levels, although the absolute maximum is much higher for the RL~planner.
When faced with strong traffic demand, the eFIFO sometimes even causes a slight decrease in flow rate for automation levels as low as \SI{20}{\percent}, compared to simple precedence rules (\SI{0}{\percent} automation).
The RL~planner, on the other hand, yields virtually monotonically rising throughput with increasing automation for any traffic demand.
Figure~\subref*{sfig:flow_rl} additionally shows the results of the legacy RL~planner that was trained in fully automated traffic for comparison.
As the deviation in flow rate is negligible, it can be concluded that the training in mixed traffic has no negative impact on the policy's peak performance.
This might be explained by the first third of the training being conducted in fully automated traffic, which enables the network to optimize for this specific case, which is supported by the independent weight matrices per edge type.

\begin{figure}
	\centering
	\includegraphics[width=\linewidth]{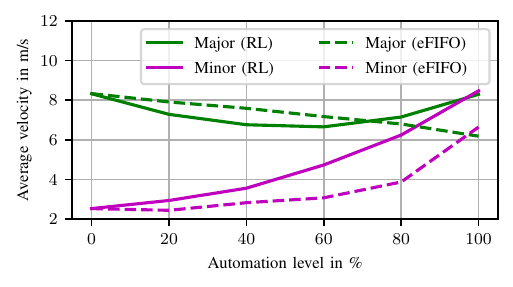}
	\caption{The average vehicle velocity depending on the vehicles' originating road and the overall level of automation. The RL~planner and baseline results are given for direct comparison.}
	\label{fig:road_velocity}
\end{figure}

Cooperative planning significantly increases the attained velocity on the minor road, as illustrated in Fig.~\ref{fig:road_velocity}.
While the RL~planner yields a monotonous increase already for low automation levels, the eFIFO requires much more of the traffic being automated to attain the same benefit.
Both approaches cause a slight velocity decrease on the major road, which is expected because the cooperative maneuvers require the prioritized vehicles to refrain from crossing the intersection unconditionally.
When using the RL~planner in fully automated traffic, this effect is compensated entirely.

\begin{figure}
	\centering
	\includegraphics[width=\linewidth]{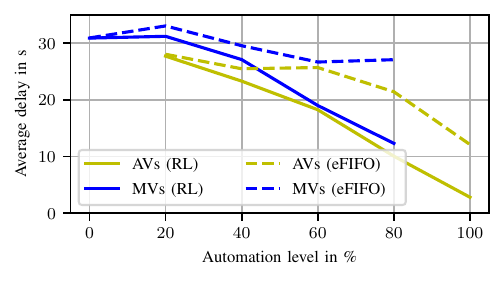}
	\caption{The average delay, as defined in \eqref{eq:delay}, induced by interaction experienced by automated and manual vehicles.}
	\label{fig:avmv_delay}
\end{figure}

Another metric of interest is the \emph{delay}, which is defined in accordance with \cite{mertens2022cooperative} for vehicle $\nu_i$ as
\begin{equation}
\mathrm{delay}(\nu_i) = \sum\limits_{k=1}^{L} 1 - \frac{v_i(k)}{v_\text{lim}(s_i(k))},
\label{eq:delay}
\end{equation}
where $L$ denotes the evaluation horizon, $v_i$ the vehicle's velocity and $v_\text{lim}$ the lane speed limit queried at the vehicles position.
Note that a delay of zero may not be attainable due to limited acceleration capabilities of the vehicles and discontinuous speed limits on the lanes.
It can be seen from Fig.~\ref{fig:avmv_delay} that an increasing share of AVs performing cooperative maneuvers does not disadvantage MVs.
They rather benefit from it, as their delay shrinks virtually alike when using the RL~planner.
This effect is less pronounced for the eFIFO, where the remaining MVs do not benefit much even in mostly automated traffic.

The collision rates in Table~\ref{tab:collisions} allow to assess to which extent cooperative planning is possible under measurement uncertainty.
Compared to the eFIFO baseline, the learned model copes significantly better with measurement uncertainties that are modeled according to Sec.~\ref{ssec:uncertainties}.
Although a collision rate of zero is not yet attained by the RL policy, the degradation due to noise is lower than for the rule-based eFIFO.
A further reduction of the collision rate is particularly challenging for a reinforcement learning setup that relies on implicit feedback through the reward signal.
In practice, these few remaining collision cases shall not cause serious problems, because a cooperative plan would be subject to feasibility checks before being executed.

\begin{table}
	\caption{Collision rates with and without measurement noise}
	\label{tab:collisions}
	\centering
	\begin{tabular}{lrr}
		\toprule
		\textbf{Measurement noise}            & \textbf{RL planner}   & \textbf{eFIFO scheme} \\ \toprule
		None                                  & \SI{0.0829}{\percent} & \SI{0.0276}{\percent} \\ \midrule
		Acc. to Sec.~\ref{ssec:uncertainties} & \SI{0.0910}{\percent} & \SI{0.3757}{\percent} \\ \bottomrule
	\end{tabular}
\end{table}

\section{Conclusion}
\label{sec:conclusion}

This work presented a novel machine learning based cooperative planning scheme for mixed traffic at urban intersections.
By training an RL policy in a simulated environment, we evaded the need for large amounts of training data, which is unavailable for cooperative maneuvers.
The proposed graph-based scene representation considers the inherent uncertainty in human-driven vehicles.
Evaluation of the RL~planner revealed a clear benefit in flow rate and reduced delays for an increasing share of AVs in traffic.
We showed that our method outperforms the eFIFO scheme for mixed traffic and is robust to measurement uncertainties.

In future works, we plan to shrink the gap to real-world application further by integrating dedicated motion planning algorithms and deploying our approach to a real testing vehicle.

\balance
\bibliographystyle{IEEEtran}
\bibliography{references}

\end{document}